\documentclass[a4paper,twoside]{article}

\usepackage{epsfig}
\usepackage{subcaption}
\usepackage{calc}
\usepackage{amssymb}
\usepackage{amstext}
\usepackage{amsmath}
\usepackage{amsthm}
\usepackage{multicol}
\usepackage{pslatex}
\usepackage{apalike}
\usepackage{algorithm2e}
\usepackage[bottom]{footmisc}

\usepackage{hyperref} 
\usepackage[capitalize,noabbrev]{cleveref} 
\usepackage{siunitx} 
\usepackage{booktabs} 
\usepackage{multirow} 
\usepackage[english=american]{csquotes} 

\usepackage{SCITEPRESS}

\newcommand{\mnist}{\mbox{MNIST}}
\newcommand{\cifarTen}{\mbox{CIFAR-10}}
\newcommand{\cifarHundred}{\mbox{CIFAR-100}}

\newcommand{\openinterval}[2]{\ensuremath{\left(#1, #2 \right)}}

\begin{document}

\title{A Convexity-dependent Two-Phase Training Algorithm for Deep Neural Networks}

\author{%
  \authorname{%
    Tomas Hrycej\sup{1}\sup{\dagger},
    Bernhard Bermeitinger\sup{2}\sup{\dagger},
    Massimo Pavone\sup{1}\sup{\ddagger},
    Götz-Henrik Wiegand\sup{1}\sup{\dagger},
    and Siegfried Handschuh\sup{1}\sup{\dagger}
  }
  \affiliation{%
    \sup{1}Institute of Computer Science, University of St.Gallen
    (HSG), St.Gallen, Switzerland
  }
  \affiliation{%
    \sup{2}Institute of Computer Science in Vorarlberg, University of
    St.Gallen (HSG), Dornbirn, Austria
  }
  \email{%
    \sup{\dagger}firstname.lastname@unisg.ch,
    \sup{\ddagger}firstname.lastname@student.unisg.ch
  }
}


\abstract{%
  The key task of machine learning is to minimize the loss function that measures the model fit to the training data.
  The numerical methods to do this efficiently depend on the properties of the loss function.
  The most decisive among these properties is the \emph{convexity} or \emph{non-convexity} of the loss function.
  The fact that the loss function can have, and frequently has, non-convex regions has led to a widespread commitment to non-convex methods such as Adam.
  However, a local minimum implies that, in some environment around it, the function is convex.
  In this environment, second-order minimizing methods such as the Conjugate Gradient (CG) give a guaranteed superlinear convergence.
  We propose a novel framework grounded in the hypothesis that loss functions in real-world tasks swap from initial non-convexity to convexity towards the optimum --- a property we leverage to design an innovative two-phase optimization algorithm.
  The presented algorithm detects the swap point by observing the gradient norm dependence on the loss.
  In these regions, non-convex (Adam) and convex (CG) algorithms are used, respectively.
  Computing experiments confirm the hypothesis that this simple convexity structure is frequent enough to be practically exploited to substantially improve convergence and accuracy.%
}
\onecolumn \maketitle \normalsize \setcounter{footnote}{0} \vfill

\section{INTRODUCTION}\label{sec:introduction}
Fitting model parameters to training data is the fundamental task of Machine Learning~(ML) with parameterized models.
The sizes of the models have experienced extraordinary growth, recently reaching hundreds of billions.
This makes clear that the efficiency of the optimization algorithm is of key importance.
The optimization consists of minimizing an appropriate loss criterion such as \emph{Categorical Cross-Entropy}~(CCE), \emph{Mean Squared Error}~(MSE), or many other variants.
These criteria are multidimensional functions of all model parameters.
From the viewpoint of solvability, there are three basic classes of unconstrained minimization tasks according to the characteristics of the minimized function:
\begin{enumerate}
  \item Convex functions
  \item Non-convex functions with a single local minimum (which is also a global minimum)
  \item Non-convex functions with multiple local minima
\end{enumerate}

\noindent Non-convex functions are frequently referred to as a single group in the ML literature.
This aggregation shadows a significant difference.
In practical terms and for typical numbers of trainable parameters of current models, global minimization of a general function with multiple local minima is infeasible (see~\cref{sec:related_work}).
By contrast, gradient descent can practically minimize non-convex functions with a single local minimum.
Every descending path will reach the minimum with certainty if it is not trapped in singularities.
For convex loss functions, the odds are even better.
The classical theory of numerical optimization provides theoretically founded algorithms with a guaranteed convergence speed, also referenced in~\cref{sec:related_work}.

From the viewpoint of this problem classification, it is well known that loss functions with popular nonlinear models can possess multiple local minima, and thus count to the last class mentioned.
Some of these minima are equivalent (such as those arising through permutations of hidden-layer units), but others may not.
So, the paradoxical situation concerning the training of nonlinear models is that methods are used that almost certainly cannot solve the problem of finding a global minimum.
The implicit assumption is that the existence of multiple minima can be neglected in the hope that the concretely obtained local minimum is sufficiently suitable for the application.
The positive experience with many excellent real models seems to justify this assumption.
What remains is distinguishing between two former basic classes: convex functions and non-convex functions with a single minimum (further referred to simply as \emph{non-convex}).

The fact that the loss functions of popular architectures are potentially non-convex has led to the widespread classification of these loss functions as non-convex.
However, from a theoretical viewpoint, the loss function is certainly convex in some environment of the local minimum.
This axiomatically results from the definition of a local minimum of any smooth function $L(x)$ by the gradient being zero:
\begin{equation}\label{eq:convexity}
  \nabla L(x) = 0
\end{equation}
and the Hessian
\begin{equation}\label{eq:convex_gradient}
  H(x) = \nabla^2 L(x)
\end{equation}
being positive definite, i.e., having positive eigenvalues.
There, convex minimization algorithms are certainly worth using.
This \emph{guaranteed} convex region can optionally be --- and frequently is --- surrounded by a non-convex region.

From this point of view, the key question for algorithm choice is where the loss function is convex and where not.
Although it is known that, in general, there may be an arbitrary patchwork of convex and non-convex subregions, a simpler, while not universally valid, assumption may exist that covers typical model architectures and application tasks.
One such assumption is formulated in~\cref{sec:convex_non-convex}.
In the next step, we will propose the appropriate optimization procedure accordingly (\cref{sec:two_phase_optimization}).
If an assumption about a typical distribution of convexity is tentatively adopted, it is crucial to check how frequently this assumption applies in the spectrum of application problems.
Although an extensive survey is not feasible due to resource limitations, experiments with a variety of typical architectures (with a focus on a Transformer and some of its simplified derivatives) are performed and reviewed to determine the validity of the assumption and the efficiency effect of optimization (\cref{sec:computing_experiments}).

\section{RELATED WORK}\label{sec:related_work}
The alleged infeasibility of minimizing functions with multiple local minima is based on algorithms available after decades of intensive research.
Heuristics, such as momentum-based extensions of the gradient method, alleviate this problem by possibly surmounting barriers between individual attractors.
Still, there is no guarantee (and also no acceptable probability) of reaching the global minimum in a finite time, since the number of attractors and boundaries between them is too large.
Similarly, methods based on annealing or relaxation~\cite{metropolis1953EquationStateCalculations,kirkpatrick1983OptimizationSimulatedAnnealing} show asymptotical convergence in probability, but the time to reach some probabilistic bounds is by far unacceptable.
Algorithms claiming complete coverage of the parameter space, like those based on Lipschitz constant bounds, or so-called clustering and Bayesian methods such as~\cite{rinnooykan1987StochasticGlobalOptimization2,mockus1997BayesianHeuristicApproach} are appropriate for small parameter set sizes less than ten.

By contrast, for non-convex functions with a single local minimum, every descending path will reach the minimum with certainty if not trapped in singularities.
Today's algorithms, such as Adam~\cite{kingma2015AdamMethodStochastic}, focus on efficiency in following the descending path.
There are convergence statements, for example, by~\cite{fotopoulos2024ReviewNonconvexOptimization,chen2022PracticalAdamNonconvexity}.
An interesting proposal for transforming a non-convex unconstrained loss function to a convex one with constraints is by~\cite{ergen2023ConvexLandscapeNeural}.
However, this approach applies only to neural networks with one hidden layer and the ReLU activation function.
A good option for covering both non-convex and convex regions would be second-order algorithms with adaptive reaction to local non-convexity, such as some variants of the \emph{Levenberg-Marquardt} algorithm~\cite{levenberg1944MethodSolutionCertain,press1992NumericalRecipes2nd}.
This algorithm is specific for least-squares minimization.
It entertains a kind of \enquote{convexity weight} of deciding between a steepest gradient step and the step towards the estimated quadratic minimum.
Unfortunately, the algorithm requires storing an estimate of the Hessian, which grows quadratically in the number of parameters, which makes it clearly infeasible for billions of parameters, even if using sparse Hessian concepts.

For convex loss functions, a numerical algorithm with a guaranteed convergence speed could be \emph{nonlinear conjugate gradient method}~\cite{fletcher1964FunctionMinimizationConjugate} and~\cite{polak1969NoteConvergenceMethodes}.
Both versions and their implementations are explained in~\cite{press1992NumericalRecipes2nd}.
They exploit the fact that convex functions can both be approximated quadratically.
This quadratic approximation has an explicit minimum whose existence can be used to approach the nonquadratic but convex function minimum iteratively, with the guarantee of superlinear convergence.

\section{CONVEX AND NON-CONVEX REGIONS OF LOSS FUNCTIONS}\label{sec:convex_non-convex}
\begin{figure*}
  \centering
  \begin{minipage}[t]{0.48\textwidth}
    \includegraphics[width=0.95\columnwidth]{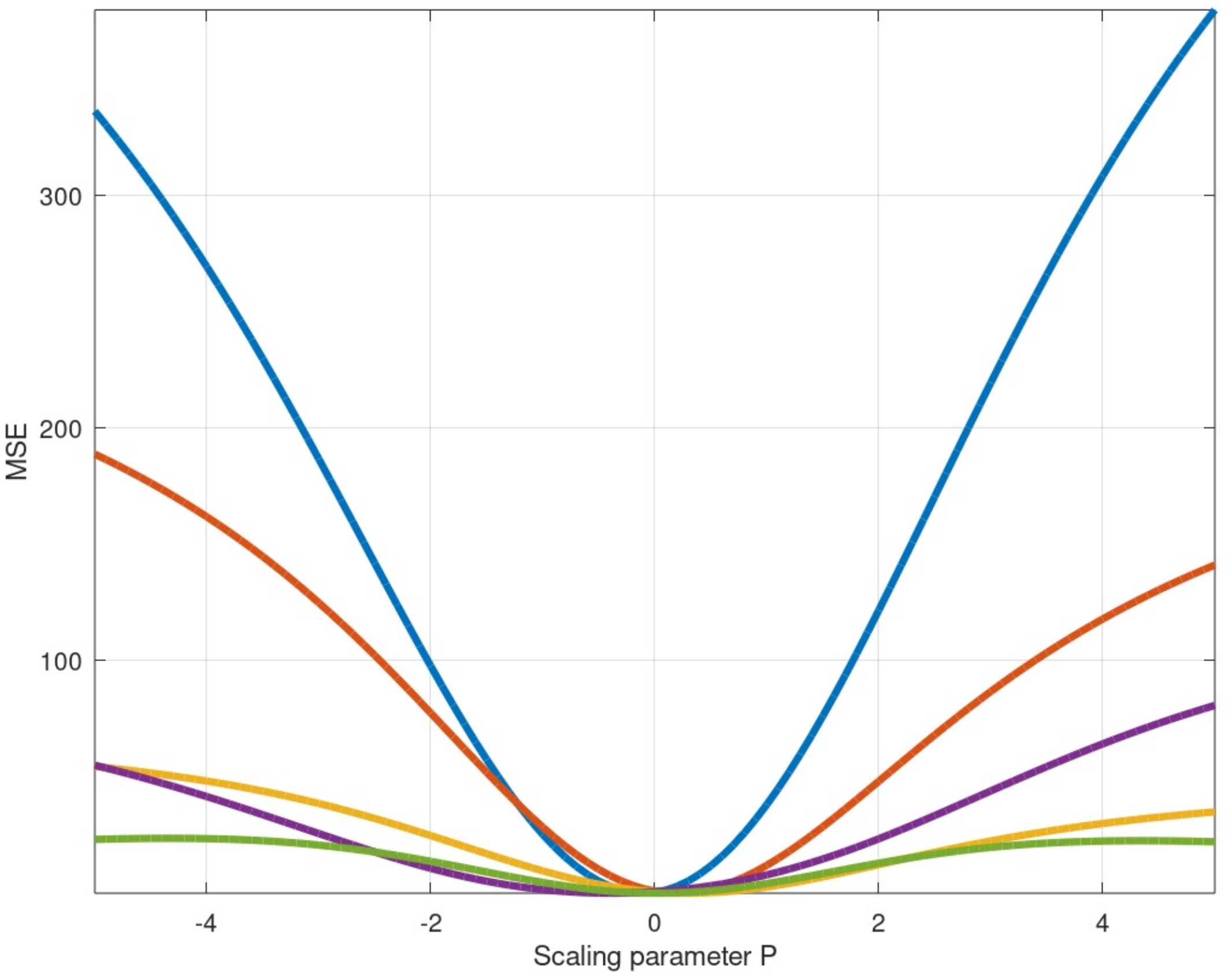}
    \caption{Loss functions of random trivial models.}\label{fig:tanh_loss}
  \end{minipage}
  \begin{minipage}[t]{0.48\textwidth}
    \includegraphics[width=0.95\columnwidth]{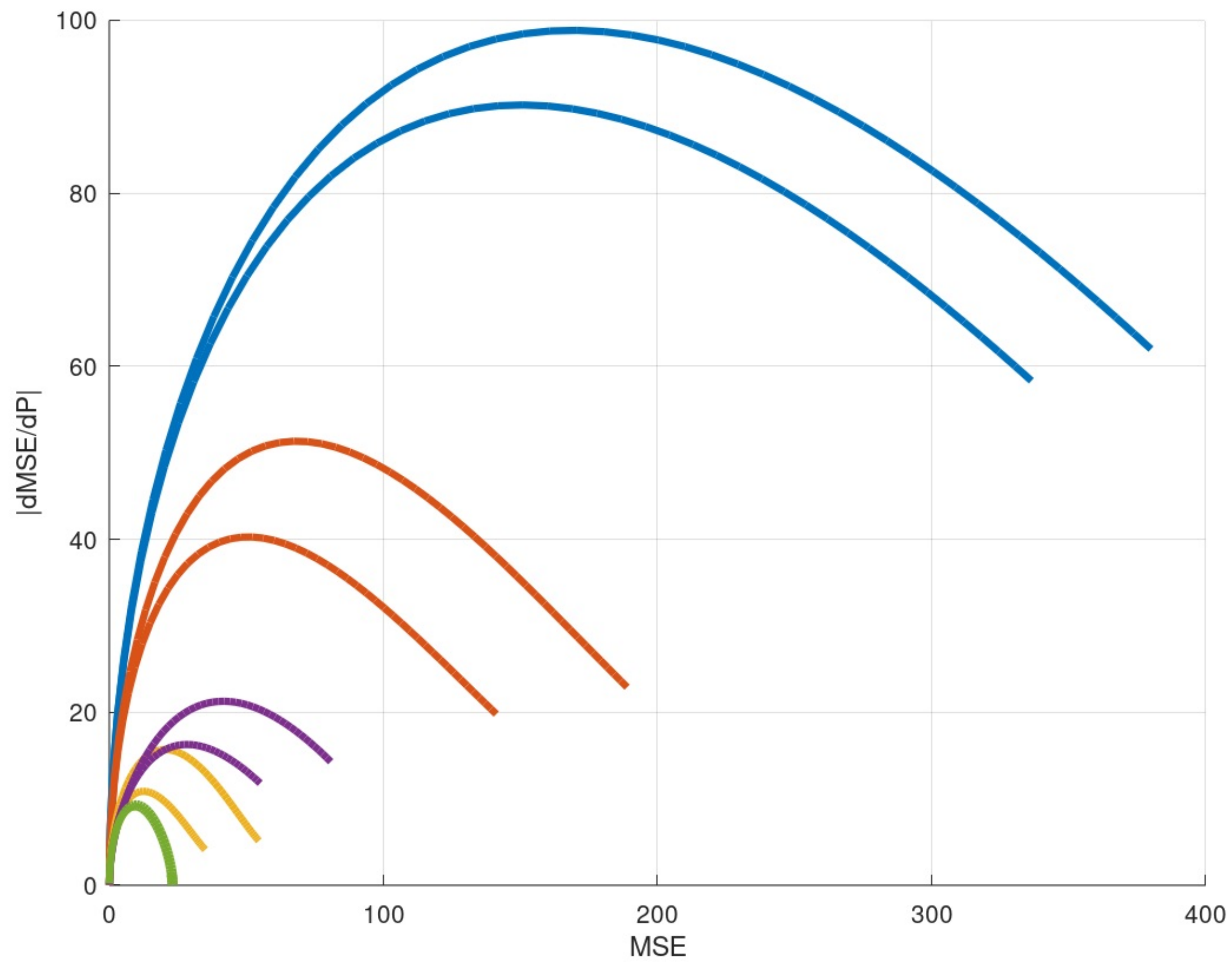}
    \caption{Dependence of the gradient norm on the loss for the random trivial models.}\label{fig:tanh_loss_grad}
  \end{minipage}
\end{figure*}

In this section, the hypothesis will be pursued that the following constellation characterizes the typical case:
There is a convex region around the minimum, surrounded by a non-convex region.
We are aware that this hypothesis will not apply to arbitrary tasks.
However, if this were frequently the case in typical applications, it could be exploited for a dedicated use of first- and second-order algorithms, respectively.

A pictorial representation of the situation is given in~\cref{fig:tanh_loss} showing the dependence of MSE on the scaling parameter $p$ for a set of five random tasks with a single nonlinear layer $\tanh$ model (with 100 units)
\begin{equation}\label{eq:tanh}
  y(x) = \sum_i \tanh{(px)}
\end{equation}
and its square loss
\begin{equation}\label{eq:tanh_loss}
  L(x) = {\big[ y(x) - r \big]}^2
\end{equation}
with reference values $r$ of the output $y$ randomly drawn from \openinterval{0}{1}.
The set is generated for randomly selected input arguments $x$ from \openinterval{-0.5}{0.5}.
Convexity around the minimum and non-convexity at margin areas can be observed.

A different view of the same five random tasks is the dependence of gradient norm on the loss, as depicted in~\cref{fig:tanh_loss_grad}.
The gradient norm is trivial in the one-dimensional case: it is the absolute value of the derivative.
During optimization, the loss on the $x$-axis decreases (from the right to the left).
The gradient norm (the $y$-axis) first increases (the non-convex region) and then decreases (the convex region) - this pattern can be observed for all five tasks.
The two branches per task correspond to the different paths to the minimum (starting at the left or at the right margin, respectively, in~\cref{fig:tanh_loss}).
It should be noted that there is no guarantee for this simple convexity pattern.
Our hypothesis is that this pattern is frequently encountered and is not universally valid.

\begin{figure}
  \centering
  \includegraphics[width=0.95\columnwidth]{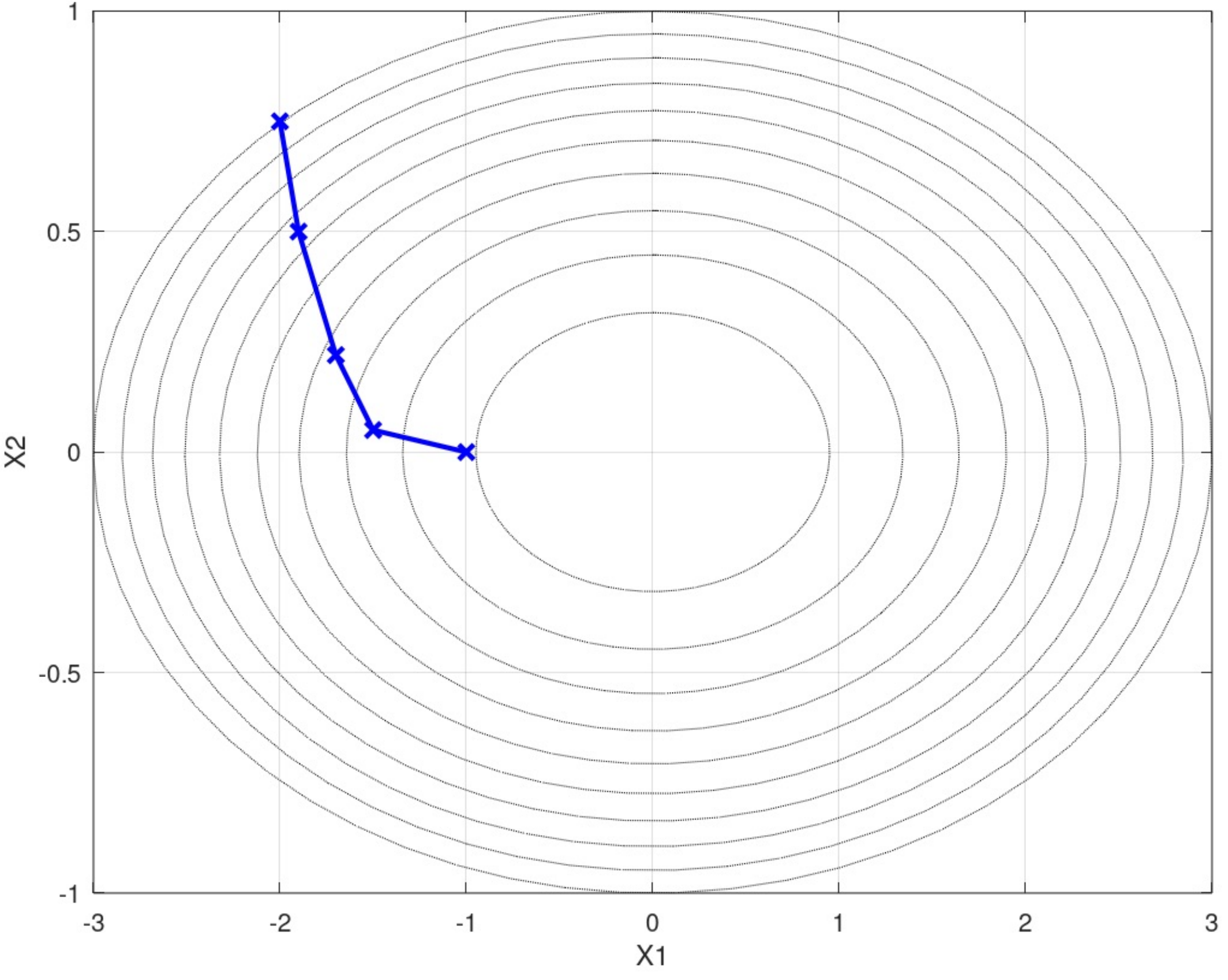}
  \caption{Gradient descent across level curves of a 2D parameter space.}\label{fig:grad_desc_2d}
\end{figure}

In the multidimensional parameter space, vertical cross-sections of a convex function are also convex so that the property of diminishing gradient norm is retained.
This is also the case for steepest gradient paths, such as that given in the 2D plot of~\cref{fig:grad_desc_2d}; the level curves become successively less dense along the path.
Of course, with an inappropriate step size, the optimization trajectory may contain segments with a temporarily increasing gradient norm if \enquote{climbing back the slope}.

\begin{figure}
  \centering
  \includegraphics[width=0.95\columnwidth]{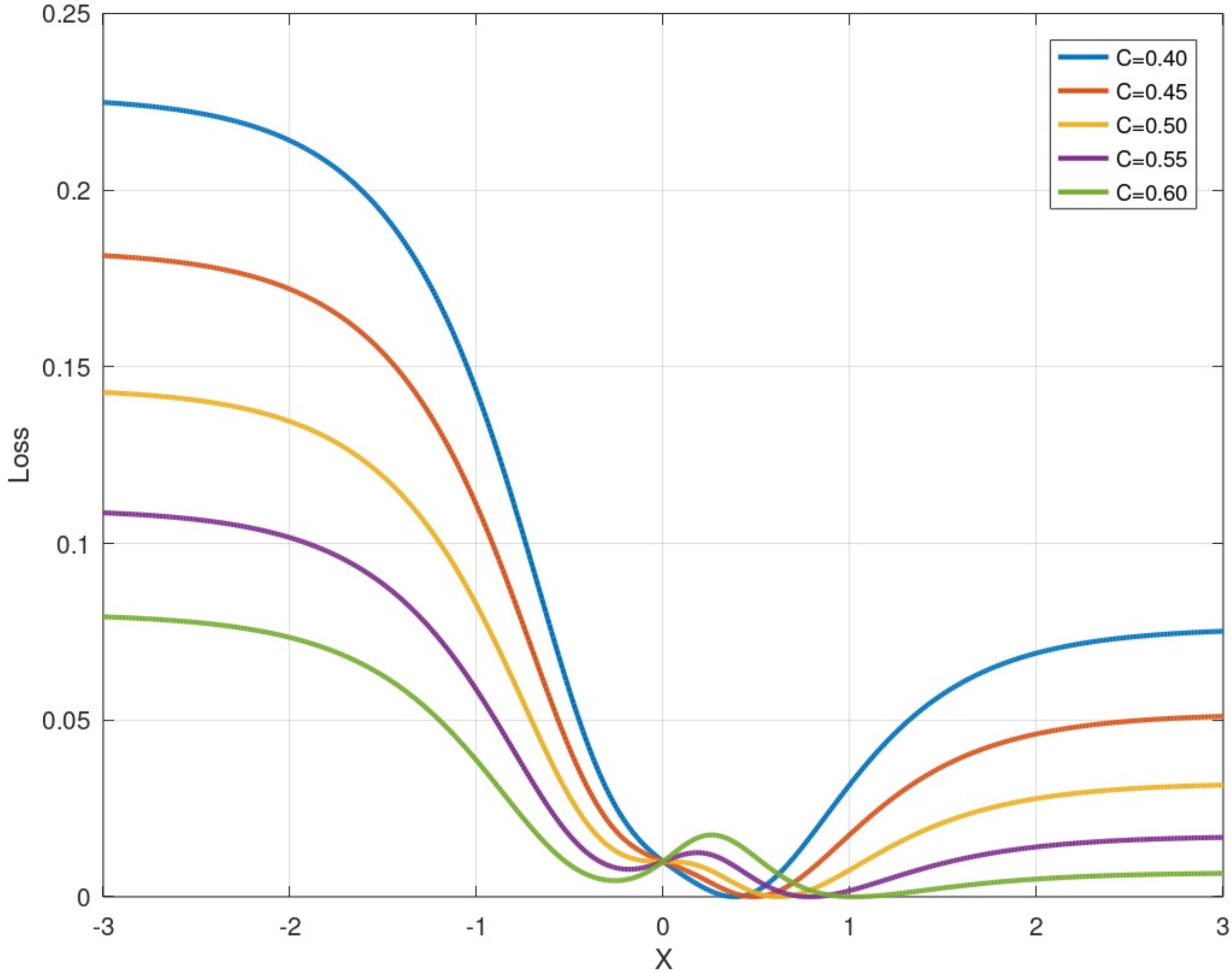}
  \caption{%
    Loss function of a trivial model with two $\tanh$ layers, with various weights $C$.
  }\label{fig:tanh2_loss}
\end{figure}

Real-world models are incomparably more complex.
Theoretically, the patterns of non-convex regions may be alternating with intermediary convex segments, forming an arbitrary patchwork.
This pitfall is analogous to those loss functions that can (and almost certainly) have multiple local minima, as mentioned in~\cref{sec:introduction}.
Alternating convex and non-convex regions are, in fact, an \emph{early stage of arising multiple local minima}.
Observing a trivial two-layer network with the hidden layer
\begin{equation}\label{eq:tanh2a}
  h(x) = \tanh{\left( x \right)}
\end{equation}
and output layer
\begin{equation}\label{eq:tanh2b}
  y(x)
  =
  \tanh{\left( h \left( x \right) \right)}
  +
  C \tanh{\left( -2 h \left( x \right) \right)}
\end{equation}
with a varying weight $C$, the loss function from \cref{eq:tanh_loss} will look like those in \cref{fig:tanh2_loss}.
For $C=0.40$, there is a single inner convex region.
For $C=0.45$ and $C=0.50$, additional local convex regions (followed by a non-convex one) arise on the left slope.
For $C=0.55$ and $C=0.60$, these convex regions convert to additional local minima.

However, the risk associated with an incorrect assumption about convexity is not as severe as in the case of one or multiple local minima.
Using convex algorithms in a non-convex region is not disastrous: the only consequence is the loss of guarantee of superlinear convergence speed.
A similarly moderate effect is using non-convex algorithms (e.g., Adam) in a convex setting.
In this sense, it can nothing but be useful to commit to an optimistic assumption that
\begin{itemize}
  \item
    the initial, usually random, parameter state is located in a non-convex region with a growing gradient norm and
  \item
    the boundary to the convex region is reached after the gradient norm decreases systematically
\end{itemize}
as in~\cref{fig:tanh_loss}.
The expectation of a multidimensional loss function behaving approximately this way is not unreasonable, although not guaranteed.
We will base our following considerations on this assumption and check how far they are encountered in real-world problems.
Then, it is possible to approximately identify the extension of non-convex and convex regions in algorithmic terms.
If the optimization algorithm is such that it produces a strictly decreasing loss (such as algorithms using line search), the entry to the convex region can be identified solely by detecting the point where the gradient norm starts its decrease.
If loss fluctuations on the optimization path appear as in stochastic gradient methods, it is more reliable to observe the dependence of the gradient norm on the loss.
In reality, both criteria may be disturbed by a zigzag optimization path in which the descent across loss-level curves does not always occur consistently.
Then, some smoothing of the gradient norm curve has to be performed.

\section{TWO-PHASE OPTIMIZATION}\label{sec:two_phase_optimization}
The basic hypothesis is as follows.
Second-order numerical optimization methods, such as the \emph{Conjugate Gradient} (CG) algorithm, can be assumed to be more efficient than first-order methods within the convex region.
By contrast, the former methods offer no particular benefits in the non-convex regions.
Then, sophisticated first-order methods (such as Adam) may be substantially more economical in their computational requirements because they use batch gradients.
To do this, it is crucial to separate both regions during optimization.
Following the principles presented in~\cref{sec:convex_non-convex}, the development of the gradient norm and its relationship with the loss currently attained can be used to detect the separating boundary.

The preceding ideas about gradient regions suggest a two-phase optimization formulated in~\cref{alg:two_phase_algorithm}.
Consistently with the hypothesis of non-convex and convex regions following the simple pattern depicted in \cref{sec:convex_non-convex}, it is necessary to identify the point where the non-convex region transitions to the convex one.
This point can be recognized with the help of an increasing or decreasing gradient norm.
The swap point between the non-convex and convex regions is thus defined as the point where the increase changes to the decrease.

However, in practical terms,  the computed gradient norm is contaminated by imprecision.
In particular, the Adam algorithm with its batch-wise precessing delivers fluctuating values (as consecutive batches are different and thus show discontinuities).
Gradient norms of the CG algorithm are nearly continuous, except for fluctuations caused by tolerances in the stopping rule of the line search.
(This can be observed in \cref{fig:plot_loss_grad}.)

This is why a practical rule to identify the swap point consists in setting a tolerance: a predefined gradient norm level below its peak value (here:~0.9).

The Adam algorithm was used for the first phase and CG with \emph{golden line search}~\cite{press1992NumericalRecipes2nd} for the second phase.

\begin{algorithm}[!h]
  \caption{%
    Two-Phase Algorithm to switch from Adam to CG when the gradient norm peak has reached.
  Model and data are left out for brevity.\label{alg:two_phase_algorithm}}
  \KwData{nbEpochs $> 1$}
  $adam \gets true$\;
  $gnmax \gets 0$\;
  $gnfact \gets 0.9$\;

  \For{epoch $\leftarrow 1$ \KwTo nbEpochs}{
    \eIf{$adam$}{
      \textsc{Adam()}\;
      $gn \gets $ \textsc{GetGradientNorm()}\;
      $gnmax \gets $ {\bfseries max}$(gn, gnmax)$\;
      $adam \gets gn > (gnmax * gnfact)$\;
    }{
      \textsc{ConjugateGradient()}\;
    }
  }
\end{algorithm}

CG has no meta-parameters except for defining a \enquote{zero} gradient norm and a tolerance for terminating the line search.
In contrast, some tuning of Adam's meta-parameters is necessary to achieve good performance.
The batch size is of particular importance.
Some researchers argue that small batches exhibit lower losses for training and validation sets, e.g.,~\cite{keskar2017LargeBatchTrainingDeep,li2014EfficientMinibatchTraining,chen2022PracticalAdamNonconvexity}.
Consistent with this finding, in our experiments, batches greater than \num{512} elements have shown deteriorating performance (only integer powers of two have been tested).
The convergence was very slow for batches exceeding \num{2048} elements (for even larger batches, even hardly discernible).
However, batches smaller than \num{512} were also inferior.
The performance of a batch size of \num{512} was good and robust for various variants of the models and has been used in further experiments.
This size has, of course, only an experimental validity for the given datasets and models.

Whether this two-phase optimization is superior to conventional algorithms depends on the extension of the convex region.
In general, this extension is not known.
Theoretically, it might be too small for switching the algorithm to be profitable.
In contrast, optimally converging algorithms may bring essential benefits in optimum quality and convergence speed.
The alternative that prevails can only be investigated empirically.

\section{COMPUTING EXPERIMENTS}\label{sec:computing_experiments}
Empirical support for a hypothesis must always be viewed with skepticism.
Nevertheless, many statements about nonlinear models cannot be made in an ultimate theoretical way, making the resort to empirical investigation inevitable.
Doubts about the validity will arise if the experimental settings do not represent the application domain.
In today's world of very large models, scaling is difficult to cover, as most single experiments are not feasible with the means of many research institutions.
We have focused on another aspect of particular relevance to the shape of the loss function and, thus, to the relationship between convex and non-convex regions: the variety of model architectures.
As the most relevant model family based on transformers, a set of reduced transformer architectures, in addition to the full transformer, has been investigated.
Furthermore, a different architecture has been used: the convolutional network \emph{VGG5} (analogous to VGG architectures but with only five weight layers~\cite{simonyan2015VeryDeepConvolutional}).
If the results are consistent with this set of architectures, the expectation that this will frequently be the case in practice is justified.
The loss criterion has been the mean squared error (MSE) in all cases.

\begin{figure}
  \centering
  \includegraphics[width=\columnwidth]{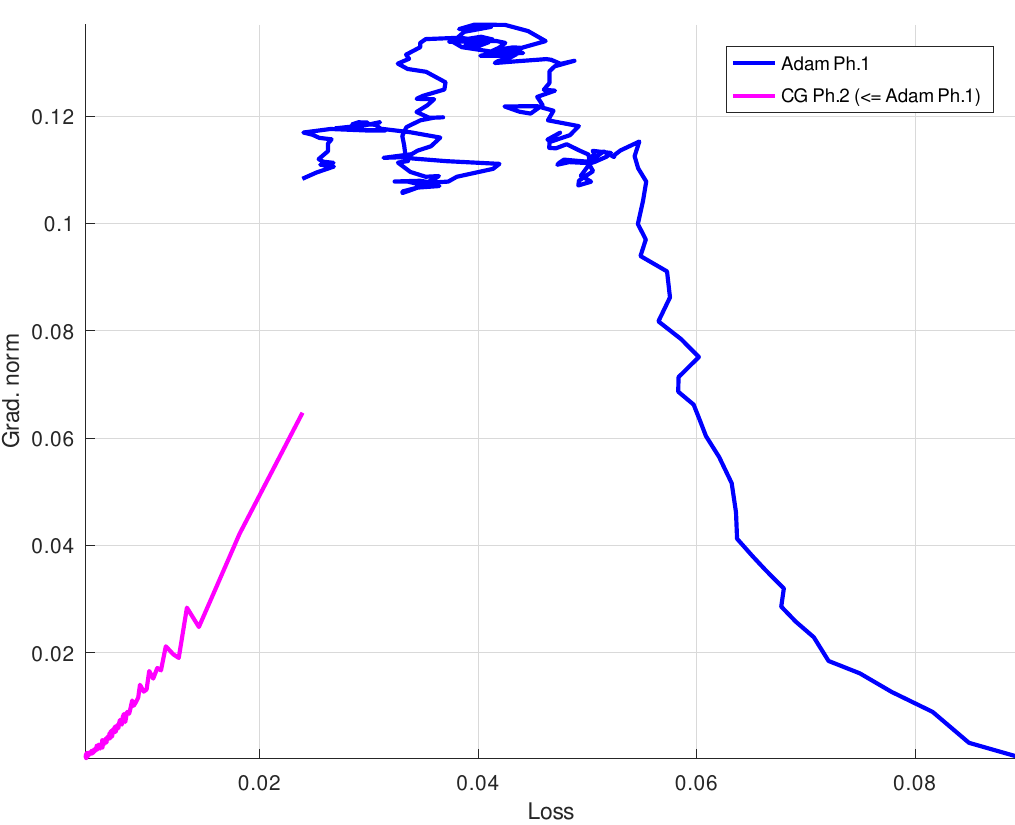}
  \caption{%
    Empirical dependence of the gradient norm on the loss, indicated here on the dataset \cifarTen{} and a ViT architecture.
    The training starts at the right side with a larger loss, decreases to the left, and decreases quickly after switching from the Adam optimizer to CG\@.
  }\label{fig:plot_loss_grad}
\end{figure}

The first series of experiments examined small variants of the \emph{Vision Transformer (ViT)} architecture~\cite{dosovitskiy2021ImageWorth16x16}.
These reduced variants consist of 3 consecutive transformer encoder layers with each 4 attention heads and a model size (embedding size) of \num{64}, in the reduced forms investigated in~\cite{bermeitinger2024ReducingTransformerArchitecture}:

\begin{itemize}
  \item
    \emph{vit-mlp}:
    a complete ViT variant with multi-head attention and multi-layer perceptron (MLP)
    The MLP is the typical two-layer neural network with one nonlinear layer with the number of units set to 4 times the model size (here: 256 units) and the activation function \emph{gelu}, followed by a linear layer to reduce the dimensions back to 64.
  \item
    \emph{vit-nomlp}:
    a variant without the MLP, thus saving many of the original model's parameters
  \item
    \emph{vit-nomlp-wkewq}:
    a variant without the MLP and additionally using a symmetric similarity measure, using the same matrix for keys and queries
  \item
    \emph{vit-nomlp-wkewq-wvwo}:
    a minimal variant additionally omitting value processing matrices $W_v$ and $W_o$
\end{itemize}
All experiments were performed with well-known datasets \cifarTen{}, \cifarHundred{}~\cite{krizhevsky2009LearningMultipleLayers}, and \mnist{}~\cite{lecun1998GradientbasedLearningApplied}.
Every experiment consists of comparing
\begin{enumerate}
  \item
    the baseline loss optimization with Adam over \num{1000} epochs (\num{700} for the MLP variants);
  \item
    an initial optimization with Adam for \num{300} epochs (\num{210} with MLP);
    followed by a further optimization with CG over \num{700} epochs (\num{490} with MLP), using the result of the preceding Adam optimization as an initial parameter state.
\end{enumerate}
All variants have shown a qualitatively similar course of the epoch-wise gradient norm.
The full ViT version, including the MLP, is shown for illustration.
\Cref{fig:plot_loss_grad} shows the gradient norm in dependence on the loss (analogy to \cref{fig:tanh_loss_grad}).
The $x$-axis contains the loss values, the $y$-axis the gradient norm.
Since the loss decreases during optimization, the training progresses from right to left along this axis.
The gradient norm values are growing from high loss values (right margin of the $x$-axis) towards lower ones.
This corresponds to the non-convex region, over which optimization takes place with the help of the Adam algorithm.
A turning point can be observed at the loss value of around \num{0.04}: the gradient norm starts to decrease.
This is qualitatively analogous to the artificial example of~\cref{fig:tanh_loss_grad} and demonstrates the entry into a convex region.
Because of this convexity, the second-order CG is used after this turning point.
This phase corresponds to the magenta curve in \cref{fig:plot_loss_grad}.

\begin{figure}
  \centering
  \includegraphics[width=\columnwidth]{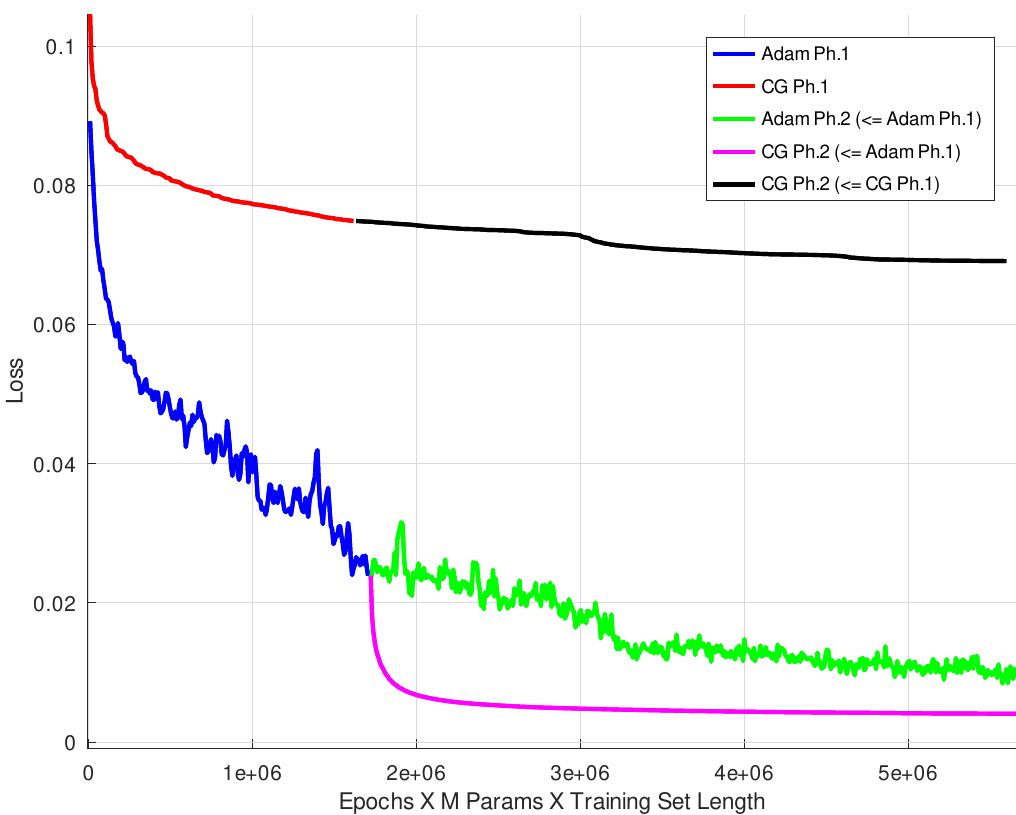}
  \caption{%
    Empirical loss function development with alternative algorithm sequences on the dataset \cifarTen{} and a ViT architecture.
    The most effective strategy is the two-phase training with Adam (blue) and CG (magenta).
    For comparison, the green line shows continuation of the Adam phase, while the red and black lines show the training purely done with CG\@.
  }\label{fig:plot_loss_fun}
\end{figure}

In~\cref{fig:plot_loss_fun}, the convergence of the loss along a magnitude approximately proportional to MFLOPS is depicted.
The blue curve shows the first phase of using Adam and the green curve shows its continuation (corresponding to using Adam in a typical way).
The second phase loss of CG (magenta curve) decreases considerably faster than its Adam counterpart (green curve).
The traditional Adam optimization over all \num{700} epochs (the blue curve and its continuation by the green curve) is visibly inferior to the convergence of the two-phase algorithm (blue and magenta curve).

The advantage of the two-phase algorithm remains substantial, even considering additional forward passes per epoch spent by line search of CG\@.
For comparison, using CG in both phases, the loss is depicted by the red and black curves.

\begin{table*}[htb!]
  \vspace{0.2cm}
  \caption{%
    Final results (loss and accuracy for the training and validation split) from the experiments on the three datasets \mnist{}, \cifarTen{}, and \cifarHundred{} for different variants of ViT and VGG5.
    The algorithm column indicates the conventional training with \textit{Adam} or the proposed second-phase training \textit{Adam+CG} using the conjugate gradient optimization method.
  }\label{tab:results}
  \centering
  \begin{tabular}{
      c 
      l 
      l 
      S[table-format=1.4, round-mode=places, round-precision=4]
      S[table-format=1.3, round-mode=places, round-precision=3]
      S[table-format=1.4, round-mode=places, round-precision=4]
      S[table-format=1.3, round-mode=places, round-precision=3]
      S[table-format=3.1, round-mode=places, round-precision=1]
    }
    \toprule
    & Model variant                & Algorithm & {Train loss}   & {Train acc.} &   {Val.\ loss} &  {Val.\ acc.} &  {$Q$} \\ \midrule
    \multirow{9}{*}{\rotatebox[origin=c]{90}{\mnist{}{}}}
    &vit-mlp               & Adam      & 0.0008 &  0.9954 & 0.0061 & 0.9652 &  3.8946 \\
    &vit-mlp               & Adam+CG   & 0.0001 &  0.9997 & 0.0044 & 0.9738 &  3.8946 \\
    &vit-nomlp             & Adam      & 0.0003 &  0.9984 & 0.0064 & 0.9628 & 10.9501 \\
    &vit-nomlp             & Adam+CG   & 0.0002 &  0.9988 & 0.0053 & 0.9685 & 10.9501 \\
    &vit-nomlp-wkewq       & Adam      & 0.0004 &  0.9977 & 0.0057 & 0.9669 & 14.1157 \\
    &vit-nomlp-wkewq       & Adam+CG   & 0.0002 &  0.9987 & 0.0048 & 0.9706 & 14.1157 \\
    &vit-nomlp-wkewq-wvwo1 & Adam      & 0.0016 &  0.9902 & 0.0073 & 0.9552 & 33.4635 \\
    &vit-nomlp-wkewq-wvwo1 & Adam+CG   & 0.0006 &  0.9963 & 0.0063 & 0.9622 & 33.4635 \\
    &vgg5-max-relu         & Adam      & 0.0001 &  0.9996 & 0.0014 & 0.9929 &  4.9183 \\
    &vgg5-max-relu         & Adam+CG   & 0.0001 &  0.9995 & 0.0011 & 0.9935 &  4.9183 \\ \midrule
    \multirow{9}{*}{\rotatebox[origin=c]{90}{\cifarTen{}}}
    &vit-mlp               & Adam      & 0.0091 &  0.9431 & 0.0997 & 0.4281 &  3.0817 \\
    &vit-mlp               & Adam+CG   & 0.0041 &  0.9703 & 0.0991 & 0.4354 &  3.0817 \\
    &vit-nomlp             & Adam      & 0.0290 &  0.8188 & 0.0981 & 0.4278 &  7.9383 \\
    &vit-nomlp             & Adam+CG   & 0.0175 &  0.8905 & 0.0982 & 0.4437 &  7.9383 \\
    &vit-nomlp-wkewq       & Adam      & 0.0386 &  0.7441 & 0.0889 & 0.4413 &  9.8623 \\
    &vit-nomlp-wkewq       & Adam+CG   & 0.0270 &  0.8328 & 0.0881 & 0.4610 &  9.8623 \\
    &vit-nomlp-wkewq-wvwo1 & Adam      & 0.0567 &  0.5746 & 0.0775 & 0.4136 & 19.1410 \\
    &vit-nomlp-wkewq-wvwo1 & Adam+CG   & 0.0527 &  0.6119 & 0.0738 & 0.4357 & 19.1410 \\
    &vgg5-max-relu         & Adam      & 0.0059 &  0.9672 & 0.0531 & 0.7102 &  4.0793 \\
    &vgg5-max-relu         & Adam+CG   & 0.0047 &  0.9691 & 0.0491 & 0.7187 &  4.0793 \\ \midrule
    \multirow{9}{*}{\rotatebox[origin=c]{90}{\cifarHundred{}}}
    &vit-mlp               & Adam      & 0.0041 &  0.7055 & 0.0128 & 0.1545 &  29.7442 \\
    &vit-mlp               & Adam+CG   & 0.0028 &  0.7579 & 0.0134 & 0.1505 &  29.7442 \\
    &vit-nomlp             & Adam      & 0.0062 &  0.4784 & 0.0112 & 0.1656 &  72.6364 \\
    &vit-nomlp             & Adam+CG   & 0.0053 &  0.5343 & 0.0116 & 0.1649 &  72.6364 \\
    &vit-nomlp-wkewq       & Adam      & 0.0069 &  0.4246 & 0.0108 & 0.1739 &  88.4205 \\
    &vit-nomlp-wkewq       & Adam+CG   & 0.0059 &  0.4874 & 0.0109 & 0.1760 &  88.4205 \\
    &vit-nomlp-wkewq-wvwo1 & Adam      & 0.0082 &  0.2908 & 0.0099 & 0.1569 & 156.3868 \\
    &vit-nomlp-wkewq-wvwo1 & Adam+CG   & 0.0078 &  0.3261 & 0.0097 & 0.1637 & 156.3868 \\
    &vgg5-max-relu         & Adam      & 0.0032 &  0.7553 & 0.0108 & 0.3004 &  38.9347 \\
    &vgg5-max-relu         & Adam+CG   & 0.0032 &  0.7372 & 0.0102 & 0.3205 &  38.9347 \\ \bottomrule
  \end{tabular}
\end{table*}

This pattern occurred for all investigated model variants and datasets (ViT variants and VGG5 with \cifarTen{}, \cifarHundred{}, and \mnist{}).
The sustained simplicity of this pattern was striking and somewhat unexpected.
There were no indicators for saddle points or spurious minima, which would become apparent as regions of a very small gradient norm.
Once the gradient norm peak passed, the second-order optimization path became straightforward.
The final results comparing a pure Adam training run and a two-phase Adam+CG are presented in~\cref{tab:results}.

Furthermore, in terms of performance metrics loss and accuracy, the overdetermination ratio of each benchmark candidate has been evaluated~\cite{hrycej2023MathematicalFoundationsData}:
\begin{equation}\label{eq:q_coeff}
  Q = \frac{KM}{P}
\end{equation}
with $K$ being the number of training examples, $M$ being the length of the output vector (usually equal to the number of classes) and $P$ being the number of trainable model parameters.

This formula justifies itself by ensuring that the numerator $KM$ is equal to the number of constraints to be satisfied (the reference values for all training examples).
This product must be larger than the number of trainable parameters for the system to be sufficiently determined.
Otherwise, there are infinite solutions, most of which do not generalize.
This is equivalent to the requirement for the overdetermination ratio $Q$ to be larger than unity.
On the other hand, too large $Q$ values may explain a poor attainable performance --- the model does not have enough parameters to represent the input/output relationship.
This is the case for \cifarHundred{}.

For the evaluation of the hypothesis formulated in~\cref{sec:convex_non-convex}, only the loss values (that is, MSE) on the training set are significant since this magnitude is what is directly minimized and thus tests the efficiency of the minimization algorithm.
There, sustained superiority of the two-phase concept can be observed.

Nevertheless, the superiority can also be extended to the accuracies and the validation set measures.
The extent of the generalization gap (the performance difference between the training and the validation sets) varies greatly.
In most cases, they can be explained by the overdetermination ratio: its large values coincide with a small training gap.
This does not apply across model groups; VGG5 generalizes better than ViT for given model architectures.

Most models used here do not reach peak performances reached by optimally tuned models for image classification.
They are typically substantially smaller to allow for the experiment series with a sufficient number of epochs.
Low epoch numbers would bring about the risk of staying in the initial non-convex region without approaching the genuine minimum.

\section{CONCLUSION}\label{sec:conclusion}
Our empirical results strongly support the hypothesis that loss functions exhibit a predictable convexity structure proceeding from the initial non-convexity towards final convexity, enabling targeted optimization strategies that outperform conventional methods.
Initial weight parameters (small random values) fall into the non-convex region, while a broad environment of loss minimum is convex.
The validity of this hypothesis can be observed in the development of the gradient norm in dependence on the instantaneous loss: a norm growing with decreasing loss indicates non-convexity, while a shrinking norm suggests convexity.

This can be exploited to identify the swap point (gradient norm peak) between both.
Then, an efficient non-convex algorithm such as Adam can be applied in the initial non-convex phase, and a fast second-order algorithm such as CG with guaranteed superlinear convergence can be used in the second phase.
A set of benchmarks has been used to test the validity of the hypothesis and the subsequent efficiency of this optimization scheme.
Although they are relatively small to remain feasible with given computing resources, they cover relevant variants of the \emph{ViT} architecture that can be expected to impact convexity properties: using or not using an MLP, defining the similarity in the attention mechanism symmetrically or asymmetrically, and putting the value vectors of embeddings in a compressed or uncompressed form (matrices $W_v$ and $W_o$).
A completely different architecture, the convolutional network \emph{VGG5}, has also been tested.

The results have been surprisingly unambiguous.
All variants exhibited the same pattern of the gradient norm increasing towards a swap point and decreasing after it.
The final losses with a two-phase algorithm have always been better than those with a single algorithm (Adam).
CG alone did not perform well in the initial non-convex phase, which caused a considerable lag so that the convex region was not attained.
The same is true with a single exception for \cifarHundred{}.
An analogical behavior can be observed for the performance of the validation set, which has been admittedly relatively poor for \cifarHundred{} because of the excessive overdetermination with given models --- the parameter sets seem to have been insufficient for image classification with 100 classes.
The top-5 accuracy on this dataset was more acceptable, over \SI{50}{\percent}.

Of course, it must be questioned how far this empirical finding can be generalized to arbitrary architectures, mainly to large models.
One of the very difficult questions is the convexity structure of loss functions with arbitrary models or even with a model class relevant to practice.
However, it is essential to note that there is no particular risk when using the two-phase method.
Gradient norms can be automatically monitored and deviations from the hypothesis can be identified.
If there is evidence against a single gradient norm peak corresponding to the swap point, a non-convex method can be used to continue as a safe resort.
If the hypothesis can be confirmed, there is an almost certain reward in convergence speed and accuracy.

Nevertheless, the next goal of our work is to verify the hypothesis on a large text-based model.

\bibliographystyle{apalike}
{\small \bibliography{references}}

\end{document}